\documentclass[11pt,a4paper]{article}
\usepackage[hyperindex,breaklinks]{hyperref}
\usepackage[hyperref]{emnlp-ijcnlp-2019}
\usepackage{times}
\usepackage{latexsym}
\usepackage{booktabs}
\usepackage{soul}
\usepackage{color}

\usepackage{amssymb}
\usepackage{url}
\usepackage{xcolor}
\usepackage{soul}
\usepackage{amsmath}
\usepackage{soulpos}
\usepackage{enumitem}
\usepackage[font={small}]{caption}
\usepackage{multirow}
\usepackage{graphicx}
\usepackage{mathtools}

\aclfinalcopy % Uncomment this line for the final submission

\newenvironment{fontppl}{\fontfamily{ppl}\selectfont}{\par} % Palatino

\definecolor{color_blind_friendly_beige}{RGB}{255,190,106}
\definecolor{color_blind_friendly_green}{RGB}{64,176,166}

\DeclareMathOperator*{\argmax}{arg\,max\,}
\newcommand\numberthis{\addtocounter{equation}{1}\tag{\theequation}}
\newcommand{\hlred}[1]{{\sethlcolor{color_blind_friendly_green}\hl{#1}}}
\newcommand{\hlgreen}[1]{{\sethlcolor{color_blind_friendly_beige}\hl{#1}}}
\ulposdef{\ulnumaux}{%
   $\underset{\saveulnum}{\rule[-.7ex]{\ulwidth}{.4pt}}$}

\newcommand{\ulnum}[2]{%
  \def\saveulnum{#1}%
  \ulnumaux{#2}}

\title{Towards Annotating and Creating Sub-Sentence Summary Highlights}

\author{Kristjan Arumae$^\spadesuit$, \, Parminder Bhatia$^\diamondsuit$, \, Fei Liu$^\spadesuit$\\[0.8em]
$^\spadesuit$Computer Science Department, University of Central Florida\\
$^\diamondsuit$Amazon, USA\\[0.5em]
  {\tt \{arumae,parmib\}@amazon.com \quad feiliu@cs.ucf.edu}\\
}

\date{}

\begin{document}
\maketitle
\begin{abstract}
Highlighting is a powerful tool to pick out important content and emphasize. 
Creating summary highlights at the sub-sentence level is particularly desirable, because sub-sentences are more concise than whole sentences. 
They are also better suited than individual words and phrases that can potentially lead to disfluent, fragmented summaries. 
In this paper we seek to generate summary highlights by annotating summary-worthy sub-sentences and teaching classifiers to do the same.
We frame the task as jointly selecting important sentences and identifying a single most informative textual unit from each sentence. 
This formulation dramatically reduces the task complexity involved in sentence compression.
Our study provides new benchmarks and baselines for generating highlights at the sub-sentence level.
\end{abstract}

\section{Introduction}
\label{sec:intro}

Highlighting at an appropriate level of granularity is important to emphasize salient content in an unobtrusive manner.
A small collection of keywords may be insufficient to deliver the main points of an article, while highlighting whole sentences often provide superfluous information.
In domains such as newswire, scholarly publications, legal and policy documents~\cite{Kim:2010,Sadeh:2013,Hasan:2014}, people are tempted to write long and complicated sentences. 
It is particularly desirable to pick out only \emph{important sentence parts} as opposed to whole sentences.

Generating highlights at the sub-sentence level has not been thoroughly investigated in the past.
A related thread of research is extractive and compressive summarization~\cite{Daume:2002,Zajic:2007,Martins:2009,Filippova:2010,Kirkpatrick:2011,Thadani:2013,Wang:2013,Li:2013:EMNLP,Li:2014:EMNLP,Durrett:2016}.
The methods select representative sentences from source documents, then delete nonessential words and constituents to form compressed summaries. 
Nonetheless, making multiple interdependent decisions on word deletion can render summaries ungrammatical and fragmented.
In this paper, we investigate an alternative formulation that can dramatically reduce the task complexity involved in sentence compression.

We frame the task as jointly selecting representative sentences from a document and identifying a \emph{single} most informative textual unit from each sentence to create sub-sentence highlights.
This formulation is inspired by rhetorical structure theory (RST; Mann and Thompson, 1988)\nocite{Mann:1988} where sub-sentence highlights resemble the \emph{nuclei} which are text spans essential to express the writer's purpose.
The formulation also mimics human behavior on picking out important content.
If multiple parts of a sentence are important, a human uses a single stroke to highlight them all, up to the whole sentence.
If only a part of the sentence is relevant, she only picks out that particular sentence part.

Generating sub-sentence highlights is advantageous over abstraction~\cite{See:2017,Chen:2018:ACL,Gehrmann:2018,Lebanoff:2018,Celikyilmaz:2018} in several aspects.
The highlights can be overlaid on the source document, allowing them to be interpreted in context.
The number of highlights is controllable by limiting sentence selection.
In contrast, adjusting summary length in an end-to-end, abstractive system can be difficult.
Further, highlights are guaranteed to be true-to-the-original, while system abstracts can sometimes ``hallucinate'' facts and distort the original meaning.
Our contributions in this work include the following:

\begin{figure*}[t]
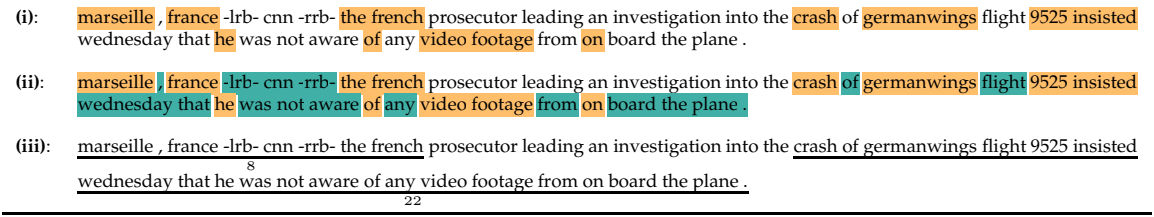

\setlength{\tabcolsep}{5pt}
\renewcommand{\arraystretch}{1}
\centering
\begin{scriptsize}
\begin{fontppl}
\begin{tabular}{ll}
\toprule
\textbf{(i)}:&\hlgreen{marseille} , \hlgreen{france} -lrb- cnn -rrb- \hlgreen{the french} prosecutor leading an investigation into the \hlgreen{crash} of \hlgreen{germanwings} flight \hlgreen{9525 insisted}\\ &wednesday that \hlgreen{he} was not aware \hlgreen{of} any \hlgreen{video footage} from \hlgreen{on} board the plane .\\\\

\textbf{(ii)}:&\hlgreen{marseille} \hlred{,} \hlgreen{france} \hlred{-lrb- cnn -rrb-} \hlgreen{the french} prosecutor leading an investigation into the \hlgreen{crash} \hlred{of} \hlgreen{germanwings} \hlred{flight} \hlgreen{9525 insisted}\\ & \hlred{wednesday that} \hlgreen{he} \hlred{was not aware} \hlgreen{of} \hlred{any} \hlgreen{video footage} \hlred{from} \hlgreen{on} \hlred{board the plane .}\\\\

\textbf{(iii)}:&\ulnum{8}{marseille , france -lrb- cnn -rrb- the french} prosecutor leading an investigation into the \ulnum{}{crash of germanwings flight 9525 insisted}\\ & \ulnum{22}{wednesday that he was not aware of any video footage from on board the plane .}\\
\bottomrule
\end{tabular}
\end{fontppl}
\end{scriptsize}
\caption{
An illustration of label smoothing. 
Words aligned to the abstract are colored \emph{orange}; gap words are colored \emph{turquoise}.
}
\label{figure:annot}
\end{figure*}

\begin{itemize}[topsep=3pt,itemsep=-1pt,leftmargin=*]

\item we introduce a new task formulation of creating sub-sentence summary highlights, then describe an annotation scheme to obtain binary sentence labels for extraction, as well as start and end indices to mark the most important textual unit of a positively labeled sentence;

\item we examine the feasibility of using neural extractive summarization with a multi-termed objective to identify summary sentences and their most informative sub-sentence units.
Our study provides new benchmarks and baselines for highlighting at the sub-sentence level. 
\end{itemize}

\section{Annotating Sub-Sentence Highlights}
\label{sec:annotation}

We propose to derive gold-standard sub-sentence highlights from human-written abstracts that often accompany the documents~\cite{Hermann:2015}. 
However, the challenge still exists, because abstracts are very loosely aligned with source documents and they contain unseen words and phrases. 
We define \emph{a summary-worthy sub-sentence unit} as the longest consecutive subsequence that contains content of the abstract. 
We obtain gold-standard labels for sub-sentence units by first establishing word alignments between the document and abstract, then smoothing word labels to generate sub-sentence labels. 

\vspace{0.05in}
\noindent\textbf{Word Alignment}\quad
The attention matrix of neural sequence-to-sequence models provides a powerful and flexible mechanism for word alignment. 
Let $S$=$\{w_i\}_{i=1}^\textsf{M}$ be a sequence of words denoting the document, and $T$=$\{w_t\}_{t=1}^\textsf{N}$ denoting the abstract. 
The attention weight $\alpha_{t,i}$ indicates the amount of attention received by the $i$-th document word in order to generate the $t$-th abstract word.
All attention values ($\boldsymbol\alpha$) can be automatically learned from parallel training data.  
After the model is trained, we identify \emph{a single document word} that receives the most attention for generating each abstract word, as denoted in Eq.~(\ref{eq:alignment}) and illustrated by Figure~\ref{figure:annot} (i).
This step produces a set of source words containing the content of the abstract but possibly with distinct word forms.\footnote{
Aligning multiple document words with a single abstract word is possible by retrieving document words whose attention weights exceed a threshold.
But the method can be data- and model-dependent, increasing the variability of alignment. 
}
\begin{align*}
w_i^{(t)} = \argmax_{i\in\textsf{M}} \alpha_{t,i} \quad \forall t
\numberthis\label{eq:alignment}
\end{align*}

\noindent\textbf{Smoothing}\quad
Our goal is to identify sub-sentence units containing content of the abstract by smoothing word labels obtained in the previous step.
We extract a single most informative textual unit from a sentence.
As a first attempt, we obtain start and end indices of sub-sentence units using heuristics, which are described as follows:
\begin{itemize}[topsep=3pt,itemsep=-1pt,leftmargin=*]

\item connecting two selected words if there is a small gap ($<$5 words) between them. 
For example, in Figure~\ref{figure:annot} (ii), the gap between ``\emph{crash}'' and ``\emph{germanwings}'' is bridged by labelling all gap words as selected;

\item the longest consecutive subsequence after filling gaps is chosen as the most important unit of the sentence. In Figure~\ref{figure:annot} (iii), we select the longest segment containing 22 words. When a tie occurs, we choose the segment appearing first;

\item creating gold-standard labels for sentences and sub-sentence units. 
If a segment is the most informative, i.e., longest subsequence of a sentence and $>$5 words, we record its start and end indices. 
If a segment is selected, its containing sentence is labelled as 1, otherwise 0.
\end{itemize}

\begin{table*}
\setlength{\tabcolsep}{13pt}
\renewcommand{\arraystretch}{1.1}
\centering
\begin{small}
\begin{tabular}{lrr|rrr|rr}
& \multicolumn{2}{c|}{\textbf{Sentences}} & \multicolumn{3}{c|}{\textbf{Gold-Standard Highlights}} & \multicolumn{2}{c}{\textbf{Human Abstracts}}\\
& \#TotalSents & \%PosSents & \#Sents & \#Tokens & \%CompR & \#Sents & \#Tokens\\
\toprule
\textbf{Train} & 5,312,010 & 24.42 & 4.51 & 51.46 & 0.47 & 3.68 & 56.47 \\
\textbf{Valid} & 211,022 & 30.85 & 4.87 & 57.11 & 0.47 & 4.00 & 62.73\\
\textbf{Test} & 182,663 & 29.63 & 4.72 & 54.47 & 0.46 & 3.79 & 59.56\\
\bottomrule
\end{tabular}
\end{small}
\caption{Data statistics are broken into three categories.  \textit{Sentences} indicate the number of total sentences as well as the rate of positive labels.  \textit{Gold-Standard Highlights} reflect document-level details of our new ground truth labels.  Compression rate (``CompR'') indicates the percentage of a positive labeled sentence was covered by the segment.  Finally \textit{Human Abstracts} provides a comparison against CNN/DailyMail ground truth summaries.}
\label{table:stats}
\end{table*}

\subsection{Dataset and Statistics}
\label{sec:data}

We conduct experiments on the CNN/DM dataset released by See et al.~\shortcite{See:2017} containing news articles and human abstracts. 
We choose the pointer-generator networks described in the same work to obtain attention matrices used for word alignment.
The model was trained on the training split of CNN/DM, then applied to all train/valid/test splits to generate gold-standard sub-sentence highlights.
At test time, we compare system highlights with gold-standard highlights and human abstracts, respectively, to validate system performance.

In Table~\ref{table:stats}, we present data statistics of the gold-standard sub-sentence highlights.
We observe that gold-standard highlights and human abstracts are of comparable length in terms of tokens.
On average, 28\% of document sentences are labelled as positive.
Among these, 47\% of the words belong to gold-standard sub-sentence highlights.
In our processed dataset we retain important document level information such as original sentence placement and document ID.
We consider each document sentence as a data instance, and introduce a neural model to predict (i) a binary sentence level label, and (ii) start and end indices of a consecutive subsequence for a positive sentence.
We are particularly interested in predicting start and end indices to encourage sub-sentence segments to remain self-contained.
Finally, we leverage the document ID to re-combine model output to still generate summaries at the document level.

\begin{table*}
\setlength{\tabcolsep}{6.7pt}
\renewcommand{\arraystretch}{1.05}
\centering
\begin{small}
\begin{tabular}{lrrr|rrr|rrr}
        & \multicolumn{3}{c|}{\textbf{ROUGE-1}} & \multicolumn{3}{c|}{\textbf{ROUGE-2}} & \multicolumn{3}{c}{\textbf{ROUGE-L}}\\
        Model & P & R & F$_1$ & P & R & F$_1$ & P & R & F$_1$\\
        \toprule
        Oracle (sent.) & 36.63 & 69.52 & 46.58 & 20.24 & 37.76 & 25.55 & 25.59 & 47.84 & 32.34\\
        Oracle (segm.) & 59.71 & 50.95 & 53.82 & 34.42 & 29.60 & \textbf{31.16} & 43.23 & 36.89 & 38.95\\
        \midrule
        Pointer Gen.{\scriptsize~\cite{See:2017}} & -- & -- & 39.53 & -- & -- &  17.28 & -- & -- &  36.38\\
        QASumm+NER{\scriptsize~\cite{Arumae:2019}} & -- & -- & 25.89 & -- & -- & 11.65 & -- & -- & 22.06\\
        \midrule
        \multirow{4}{*}{\rotatebox{90}{\textsc{Abstract}}} \quad Sent & 30.91 & 48.61 & 34.84 & 13.31 & 21.40 & 15.09 & 20.14 & 31.44 & 22.55 \\
        \quad\quad Sent + posit.  & 31.31 & 56.53 & 37.72 & 14.45 & 26.70 & \textbf{17.53} & 20.51 & 37.05 & 24.63 \\
        \quad\quad Segm   & 32.58 & 44.97 & 34.73 & 13.79 & 19.36 & 14.75 & 21.36 & 29.03 & 22.51\\
        \quad\quad Segm + posit.  & 33.11 & 52.74 & 37.99 & 14.96 & 24.30 & \textbf{17.26} & 21.69 & 34.41 & 24.75\\
        \midrule
        \multirow{4}{*}{\rotatebox{90}{\textsc{Sub-Sent}}} \quad Sent  & 38.93 & 58.49 & 42.81 & 28.88 & 44.49 & 31.96 & 32.92 & 50.14 & 36.32\\
        \quad\quad Sent + posit. & 39.97 & 68.59 & {47.02} & 31.38 & 55.31 & \textbf{37.19} & 34.58 & 60.30 & {40.86}\\
        \quad\quad Segm  & 41.31 & 54.27 & 42.83 & 30.29 & 40.38 & 31.43 & 34.81 & 46.01 & 36.07\\
        \quad\quad Segm + posit. &  42.43 & 64.09 & {47.43} & 32.75 & 50.40 & \textbf{36.76} & 36.43 & 55.58 & {40.80}\\
        \bottomrule
    \end{tabular}
    \end{small}
    \caption{ROUGE results on CNN/DM test set at both sentence and sub-sentence level.  The top two rows test gold-standard sentences and sub-sentences against human abstracts.  
    Additionally we show results of an abstractive~\cite{See:2017} and an extractive summarizer~\cite{Arumae:2019} whose CNN/DM results are macro-averaged.  
    The bottom two sections showcase our models.  We report results at sentence and sub-sentence level and report those with and without $E_{\mbox{\scriptsize d-pos}}$ embeddings (\emph{+posit.}).  These results are further broken down to reflect evaluation against human abstracts and our own gold standard segments.}
\label{table:rouge}
\end{table*}

\section{Models}
\label{sec:approach}

We provide initial modeling for our data with a single state-of-the-art architecture.
The purpose is to build meaningful representations that allow for joint prediction of summary-worthy sentences and their sub-sentence units.
Our model receives an input sequence as an individualized sentence denoted as $S$=$\{w^s_i\}_{i=1}^\textsf{M}$, where $s$ denotes the sentence index in the original document.
The model learns to predict the sentence label and start/end index of a sub-sentence unit based on contextualized representations.

For each token $w^s_i$ we leverage a combined representation $E_{\mbox{\scriptsize tok}}$, $E_{\mbox{\scriptsize s-pos}}$, and $E_{\mbox{\scriptsize d-pos}}$, i.e., a token embedding, sentence level positional embedding, and a document level positional embedding.
Here \emph{s-pos} denotes the token position in a sentence, \emph{d-pos} denotes the sentence position in a document, and $E(w^s_i) \in \mathbb{R}^d$.
We justify the last embedding by noting that the sentence position within that document plays an important role since generally there is a higher probability of positive labels towards the beginning.
The final input representation is an element-wise addition of all embeddings (Eq. \eqref{equation:enc}).
This input is encoded using a bi-directional transformer \cite{NIPS2017_7181, devlin2018bert}, denoted as $\mathbf{h}$.
\begin{equation}\label{equation:enc}
    E(w^s_i) \coloneqq E_{\mbox{\scriptsize tok}}(w^s_i) + E_{\mbox{\scriptsize s-pos}}(w^s_i) + E_{\mbox{\scriptsize d-pos}}(w^s_i)    
\end{equation}

\subsection{Objectives}
We use the transformer output to generate three labels: sentence, start and end positions of the sub-sentence unit.
First we obtain the sequence representation via the \textsf{\scriptsize [CLS]} token.\footnote{\textsf{\scriptsize [CLS]} is fine-tuned as a class label for the entire sequence, and always positioned at $\mathbf{h}_1$}
We apply a linear transformation to this vector and a softmax layer to obtain a binary label for the entire sentence.

For the indexing objective we transform the encoder output, $\mathbf{h}$, to account for start and end index classification. $\mathbf{a} = \textrm{MLP}_{\mbox{\scriptsize start/end}}(\mathbf{h}) \in \mathbb{R}^{\textsf{M} \times 2}$.
Again we make use of a single linear transformation, here it is applied across the encoder temporally giving each time-step two channels.
The two channels are individually passed through a softmax layer to produce two distributions, for the start and end index.
Finally we use a combined loss term which is trained end-to-end using a cross entropy objective:
\begin{equation}
    \mathcal{L} = \lambda (\mathcal{L}_{\mbox{\scriptsize start}} + \mathcal{L}_{\mbox{\scriptsize end}}) + \mathcal{L}_{\mbox{\scriptsize sent}}.
\end{equation}
For negatively labeled sentences $\mathcal{L}_{\mbox{\scriptsize start}}$ and $\mathcal{L}_{\mbox{\scriptsize end}}$ are not utilized during training.
$\lambda$ is a coefficient balancing between two task objectives.

\subsection{Experimental Setup}

The encoder hidden state dimension is set at $768$, with $12$ layers and $12$ attention heads (BERT\textsubscript{BASE} uncased).
We utilize dropout \cite{srivastava2014dropout} with $p=0.1$, and $\lambda$ is empirically set to $0.1$.
We use Adam \cite{kingma2014adam} as our optimizer with a learning rate of $3e^{-5}$, and implement early stopping against the validation split.
Devlin et al. \shortcite{devlin2018bert} suggest that fine-tuning takes only a few epochs with large datasets.
Training was conducted on a GeForce GTX 1080 Ti GPU, and each model took at most three days to converge with a maximum epoch time of 12 hours.

At inference time we only extract start and end indices when the sentence label is positive.
Additionally if the system produced an end index occurring before the start index we ignore it and select the argmax of the distribution for end indexes which are located after the start index.

\section{Results}
\label{sec:results}

In Table~\ref{table:rouge} we report results on the CNN/DM test set evaluated by ROUGE~\cite{lin2004rouge}.
We examine to what extent our summary sentences and sub-sentence highlights, annotated using the strategy presented in \S\ref{sec:annotation}, have matched the content of human abstracts. These are the \emph{oracle} results for sentences and segments, respectively.
Despite that abstracts can contain unseen words, we observe that 70\% of the abstract words are covered by gold-standard sentences, and 51\% of abstract words are included in sub-sentence units, suggesting the effectiveness of our annotation method on capturing summary-worthy content.

We proceed by evaluating our method against state-of-the-art extractive and abstractive summarization systems.
Arumae and Liu~\shortcite{Arumae:2019} present an approach to extract summary segments using question-answering as supervision signal, assuming a high quality summary can serve as document surrogate to answer questions.
See et al.~\shortcite{See:2017} present pointer-generator networks, an abstractive summarization model and a reliable baseline for being both state-of-the-art, and also a vital tool for guiding our data creation.
We show that the performance of oracle summaries is superior to these baselines in terms of R-2, with sub-sentence highlights achieving the highest R-2 F-score of 31\%, suggesting extracting sub-sentence highlights is a promising direction moving forward.

\subsection{Modeling}
Our models are shown in the bottom two sections of Table \ref{table:rouge}.
We obtain system-predicted whole sentences (\emph{Sent}) and sub-sentence segments (\emph{Segm});
then evaluate them against both human abstracts (\textsc{Abstract}) and gold-standard highlights (\textsc{Sub-sent}).
We test the efficacy of document positional embeddings (Eq.~(\ref{equation:enc})), denoted as \emph{+posit}.

Using R-2 as a defining metric, our model outperforms or performs competitively with both the abstractive and extractive baselines.
We find that the use of document level positional embeddings is beneficial and that for both summary types, models with these embeddings have a competitive edge against those without.
Notably sub-sentence level ROUGE scores consistently outmatch sentence level values.
These results are nontrivial, as segment level modeling is highly challenging, often resulting in increased precision but drastically reduced recall~\cite{cheng2016neural}.

Our model (\emph{+posit}) positively labeled $22.27\%$ of sentences, with an average summary length of $3.54$ sentences.
The segment model crops selected sentences, exhibiting a compression ratio of $0.77$.
Comparing to gold-standard ratio of $0.47$, there is a $67.4\%$ increase, pointing to future work on highlighting sub-sentence segments.

\section{Conclusion}
\label{sec:conclusion}
We introduced a new task and dataset to study sub-sentence highlight extraction.
We have shown the dataset provides a new upper bound for evaluation metrics, and that the use of sub-sentence segments provides more concise summaries over full sentences.
Furthermore, we evaluated our data using a state-of-the-art neural architecture to  show the modeling capabilities using this data.

\section*{Acknowledgments}

We thank the anonymous reviewers for their valuable suggestions.
This research was supported in part by the National Science Foundation grant IIS-1909603.

\bibliography{kristjan,fei_highlights}
\bibliographystyle{acl_natbib}

\end{document}